\documentclass[authorversion,screen]{jair}

\RequirePackage[
  datamodel=acmdatamodel,
  style=acmauthoryear,
  backend=biber,
  giveninits=true
  ]{biblatex}

\input{preamble/packages}
\newcommand{\header}[1]{\vspace*{1mm}\noindent\textbf{#1.}}

\newcommand{\ie}{\emph{i.e.,}\xspace}

\newcommand{\eg}{\emph{e.g.,}\xspace}

\newcommand{\ignore}[1]{}

\acrodef{OOD}{out-of-distribution}
\acrodef{LLM}{large language model}

\author{Siddharth Mehrotra}
\affiliation{%
   \institution{University of Amsterdam \& Delft University of Technology}
   \country{The Netherlands}
}
\email{s.mehrotra@uva.nl}

\author{Jin Huang}
\affiliation{%
   \institution{University of Cambridge}
   \country{United Kingdom}
}   
\email{jh2642@cam.ac.uk}

\author{Xuelong Fu}
\affiliation{%
   \institution{University of Amsterdam}
   \country{The Netherlands}
}   
\email{xlongfu@outlook.com}

\author{Roel Dobbe}
\affiliation{%
   \institution{Delft University of Technology}
   \country{The Netherlands}}
\email{r.i.j.dobbe@tudelft.nl}   

\author{Clara I. Sánchez}
\affiliation{%
   \institution{University of Amsterdam}
   \country{The Netherlands}
}   
\email{c.i.sanchezgutierrez@uva.nl}

\author{Maarten de Rijke}
\affiliation{%
   \institution{University of Amsterdam}
   \country{The Netherlands}
}   
\email{m.derijke@uva.nl}

\renewcommand{\shortauthors}{Mehrotra et al.}

\setcopyright{cc}
\copyrightyear{2025}
\acmDOI{10.1613/jair.1.xxxxx}

\JAIRAE{Min-Yen Kan}
\JAIRTrack{Survey Articles} 
\acmVolume{}
\acmArticle{}
\acmMonth{10}
\acmYear{2025}

\AtBeginDocument{%
  }

\begin{document}

\title{Understanding AI Trustworthiness: A Scoping Review of AIES \& FAccT Articles}

\begin{abstract}
\textbf{Background:} Trustworthy AI serves as a foundational pillar for two major AI ethics conferences: AIES and FAccT. However, current research often adopts techno-centric approaches, focusing primarily on technical attributes such as reliability, robustness, and fairness, while overlooking the sociotechnical dimensions critical to understanding AI trustworthiness in real-world contexts.

\noindent\textbf{Objectives:} This scoping review aims to examine how the AIES and FAccT communities conceptualize, measure, and validate AI trustworthiness, identifying major gaps and opportunities for advancing a holistic understanding of trustworthy AI systems.

\noindent\textbf{Methods:} We conduct a scoping review of AIES and FAccT conference proceedings to date, systematically analyzing how trustworthiness is defined, operationalized, and applied across different research domains. Our analysis focuses on conceptualization approaches, measurement methods, verification and validation techniques, application areas, and underlying values.

\noindent\textbf{Results:} While significant progress has been made in defining technical attributes such as transparency, accountability, and robustness, our findings reveal critical gaps. Current research often predominantly emphasizes technical precision at the expense of social and ethical considerations. The sociotechnical nature of AI systems remains less explored and trustworthiness emerges as a contested concept shaped by those with the power to define it.

\noindent\textbf{Conclusions:} An interdisciplinary approach combining technical rigor with social, cultural, and institutional considerations is essential for advancing trustworthy AI. We propose actionable measures for the AI ethics community to adopt holistic frameworks that genuinely address the complex interplay between AI systems and society, ultimately promoting responsible technological development that benefits all stakeholders.
\end{abstract}

\maketitle

\acresetall

\section{Introduction}
\label{sec:intro}



The rapid advancement and widespread adoption of artificial intelligence (AI) have ushered in a new era of technological innovation, bringing both immense potential and significant challenges. As AI increasingly permeates aspects of our lives, from healthcare to criminal justice, the need for trustworthy AI has become paramount. 
\emph{Trustworthy} AI, as a concept, encompasses a multifaceted approach to AI systems that prioritizes safety, transparency, and ethical considerations for all stakeholders \cite{10.1145/3555803}. It extends beyond technical proficiency, embracing principles like reliability, fairness, explainability, and accountability. 
This has given rise to dedicated academic venues like the \emph{AAAI/ACM Conference on AI, Ethics, and Society} (AIES) and \emph{ACM Conference on Fairness, Accountability, and Transparency} (FAccT), fostering interdisciplinary discourse on AI ethics.
It is now clear that a purely technology-centric view of trustworthiness is not enough.
Trustworthy AI requires an interdisciplinary perspective that views AI systems as sociotechnical systems.

The sociotechnical nature of AI systems demands a holistic approach to trustworthiness, one that considers not only the technical aspects but also the complex interplay between AI and the broader social, cultural, and institutional contexts in which it operates. \citet{betram} lays out five elements of trustworthiness as competent, reliable, transparent, benevolent, and having ethical integrity -- and calls out to study these elements in a broader sociotechnical setting. This expanded perspective is particularly crucial for the AI ethics community which aims to bridge the gap between computer science and other disciplines in addressing AI's ethical challenges.

In a study by \citet{laufer}, when asked about what values 36 self-selecting FAccT affiliates believe FAccT scholarship should address in the near future, participants reflected on the need for broader conceptions for trustworthiness. Central to this discourse is the recognition that the concept of trustworthiness is fundamental to understanding and predicting trust levels in AI systems. It becomes imperative to critically examine how the AIES \& FAccT communities conceptualize and communicate trustworthiness, and to what ends these efforts are directed. This examination raises important questions about the commitments that trustworthy AI research in these venues signifies, or should signify, in the broader context of AI ethics and societal impact.

The study of the trustworthiness of AI systems has been a topic of interest for many years, even before the existence of the AIES \& FAccT conferences. Scholars from various disciplines have identified values that determine the attribution of trustworthiness, revealing both similarities and differences across fields. For example, in interpersonal trust, competence, predictability, benevolence, and integrity have been highlighted as crucial values \cite{lahusen2024trust}. For public institutions, the list extends to competence and reliability, procedures like transparency and accountability, and results including effectiveness and general welfare \cite{polemi2024challenges}. In the context of AI systems, properties such as reliability, robustness, safety, interpretability, explainability, fairness, transparency, and accountability have been identified as trust-relevant \cite{kaur2021requirements,lee2004trust}.

Trustworthiness has been used to refer to two sides of a coin \cite[\eg][]{lee2004trust,liao2022designing,mayer1995integrative,de2014design,schlicker2025we}. On the one hand, trustworthiness has been referred to as an objective attribute of the trustee \cite[\eg][]{zerilli2022transparency,gillis2024trust,kelp2023trustworthiness,jacovi,floridi2018ai4people}. On the other hand, trustworthiness has been referred to as a trustor’s subjective perception of a trustee’s attributes \cite[\eg][]{mayer1995integrative,schlicker2025we,lee2004trust}. Overall, trustworthiness is multifaceted, comprising several elements regardless of whether it is viewed as an inherent quality of the trusted party or as a subjective assessment made by those extending trust \cite{baer2018people,dietz2006measuring,jacovi,lee2004trust}. Therefore, by concentrating on trustworthiness, we can assess the qualities and behaviors of AI systems that contribute to their reliability, safety, and ethical alignment. 

Despite the extensive research on trustworthy AI \cite{toney}, there remains a critical gap in understanding the sociotechnical nature of these systems. Few studies have adequately addressed the complex interplay between technical capabilities and social contexts in which AI operates. This paper aims to address these gaps by conducting a comprehensive scoping review of articles published in AIES \& FAccT conferences to date. Through our analysis, we seek to answer several key questions:
\begin{enumerate}[label=(RQ\arabic*),ref={RQ\arabic*}]
    \item How is trustworthiness conceptualized in the context of AI systems within AIES \& FAccT proceedings? \label{rq:1}
    \item What methodologies are employed to measure, verify, and validate trustworthiness?\label{rq:2}
    \item Which application areas are most prominently represented?\label{rq:3}
    \item What underlying values and ethical considerations drive this body of work?\label{rq:4}
\end{enumerate}
We explicitly focus on values because an account of value embodiment in AI aids in assessing whether designed AI systems indeed embody a range of moral values \cite{mehrotraaies}, \eg those articulated by the EU High-Level Expert Group \cite{EU_AI_Ethics_Guidelines}. To this end, we examine three key dimensions: intended, embodied and realized values following the framework of \citet{van2020embedding} to do a thematic analysis on our corpus. Our motivation to use this framework is that it helps us overcome the limitations of focusing solely on intentions (which may not manifest in the system) or outcomes (which may be influenced by external factors) by emphasizing embodied values—those intentionally and successfully embedded in the system. 

Overall, our core contributions are:
\begin{enumerate}[label=(\arabic*)]
    \item A systematic analysis of how trustworthiness is conceptualized, measured, and validated within the AIES \& FAccT community.
    \item Identification of major gaps in current research, particularly regarding the sociotechnical aspects of AI systems.
    \item A critical examination of the values and ethical considerations underpinning trustworthy AI research.
    \item A discussion of the intellectual and broader impact of AIES \& FAccT conferences in studying AI trustworthiness.
\end{enumerate}


\section{Related Work}
Trust is a much-discussed topic in algorithmic decision-making, especially in the area of AI \cite{mehrotrareview}. In the development of trust, the process by which a human assesses the trustworthiness of a system, leading to their perception of trustworthiness, is crucial. 
Only with an accurate trustworthiness assessment can people base their trust on adequate expectations about a system’s capabilities and limitations and make informed decisions.
Trustworthiness of AI systems has been studied from multiple disciplines such as computer science \cite{liu2023trustworthy}, psychology \cite{schlicker2022calibrated}, public administration \cite{lahusen2024trust}, and medicine \cite{zhang2023ethics}. 
Below, we provide a background on studying trustworthiness in four subject areas, namely: computer science, law, social sciences, and humanities. These areas are common in the AIES \& FAccT conferences. AIES includes experts from various disciplines such as ethics, philosophy, economics, sociology, psychology, law, history, and politics~\cite{AIES2018}, while FAccT brings together scholars from computer science, law, social sciences, and humanities~\cite{Facct2018}. These areas serve as the foundation for our interdisciplinary approach to understanding the multifaceted nature of AI trustworthiness.

\subsection{Computer Science}
Computer science has predominantly focused on technical perspectives for trustworthy AI applications, emphasizing three characteristics, 
\begin{enumerate*}[label=(\roman*)]
\item robustness, fortifying AI models against malicious attacks such as adversarial attacks; 
\item generalization, ensuring maintaining performance on unseen \ac{OOD} data; and 
\item interpretability, improving understanding of AI model predictions~\citep{wang2023trustworthy,mucsanyi2023trustworthy}. 
\end{enumerate*}
%
%
\citet{mucsanyi2023trustworthy} list common pitfalls in evaluating trustworthy machine learning models, \eg inconsistent coding for evaluation metrics despite being the same mathematically, confounding multiple factors in method comparisons, training and test samples overlap, and lack of validation set.
Building on the foundational principles of trustworthy machine learning, recent research has begun to investigate the characteristics that apply to \acp{LLM} and mechanisms for evaluating their trustworthiness. 
\citet{liu2023trustworthy} survey key dimensions for assessing \ac{LLM} trustworthiness. These include reliability, safety, fairness, resistance to misuse, explainability and reasoning, adherence to social norms, and robustness. \citet{liu2023trustworthy} highlight that a key principle for evaluating an \ac{LLM}'s trustworthiness is the generation of proper test data across the aforementioned dimensions. Finally, \citet{toney} review similarities and differences between governments’ and researchers’ definitions and frameworks on trustworthy AI. The authors find inconsistencies between policy and research term frequencies, highlighting the different focuses of each group on trustworthy AI, distinct from our work focusing on AI trustworthiness rather than the use of the terminology within the  AIES \& FAccT community.

Studies in computer science are often tech-centred, focusing primarily on technical methodologies but overlooking socio-cultural, ethical, and legal dimensions of trustworthiness.
But achieving trustworthiness demands interdisciplinary collaboration. Thus, we consider AI systems as socio-technical and explore how the AIES \& FAccT community is deepening our insight into trustworthiness of AI.

\subsection{Social Sciences}
Social science provides a broad perspective on AI trustworthiness focusing on its societal, institutional, economic, and political implications and dynamics. E.g., \citet{dacon2023you} investigates the impact of AI trustworthiness on various aspects of society, like the environment, human society, and societal values. These implications are abstracted into three social principles: 
\begin{enumerate*}[label=(\roman*)]
\item harm prevention, which focuses on ensuring safety, security, reliability, and privacy; 
\item explicability, which emphasizes explainability and transparency of the system; and 
\item fairness, which consists of accountability and the well-being of society and environment. 
\end{enumerate*}
\citet{thiebes2021trustworthy} evaluate  AI trustworthiness frameworks and synthesize the recurring themes in five principles: 
beneficence, non-maleficence, autonomy, justice, and explicability.
They also conceptualize the \textit{DaRe4TAI} framework, which structures and directs data-driven research in trustworthy AI while addressing tensions between the five principles. Similar framework-oriented approaches are pursued by other social science scholars, as they form a crucial connection with policymaking \citep{polemi2024challenges, kusche2024possible}. We adopt a similar lens in this review as \citet{polemi2024challenges, kusche2024possible} to discuss the results of our corpus analysis with a socio-technical focus by thinking about the entire ecosystem in which our AI systems operate.

\subsection{Law}
The question of AI trustworthiness is crucial in the legal domain, where even minor inaccuracies can have significant consequences for individuals navigating the complexities of legal processes.  Studies have demonstrated the capabilities of LLMs in understanding and responding to natural language queries.  However, the tendency of LLMs to generate inaccurate and incomplete information raises serious concerns about their trustworthiness in providing legal guidance \cite{kattnig2024assessing}. E.g., \citet{tan2023chatgpt} investigate ChatGPT’s ability to provide legal information 
using several simulated cases and compared with the performance to that of 
and find that often leads laypeople to over-trust it. 
More generally looking at use of AI systems in law, \citet{steenhuis2024ai} raises important questions of exploring trustworthiness of such systems for methods of service delivery, replacement of routine legal tasks and essential legal assistance to those who might otherwise go without. Potential answers to these explorations can be derived from \citet{hagan2020legal}'s work who provides legal design testing metrics to explore trustworthiness of AI systems plus methodology and “design deliverables” based on these metrics with California state courts’ Self Help Centers \cite{hagan2018human}. Overall, AI systems hold significant promise for improving accessibility and efficiency in the legal domain but face challenges related to inaccuracies, hallucinations, and over-reliance, raising concerns about trustworthiness.

\subsection{Humanities}
The study of AI trustworthiness within the humanities emphasizes ethical, philosophical, and cultural aspects, addressing critical issues such as moral agency, responsibility, and the societal impact of AI decision-making \cite{noller2024extended, rawas2024ai, chun2023crisis}. \citet{balmer2023sociological} critically examines AI’s role in society through creative methodologies, while others explore how AI reshapes creativity, authorship, and cultural interpretation \cite{lim2018ai}. Historical and philosophical analyses \citep[\eg][]{coeckelbergh2022political} provide valuable context for understanding contemporary AI technologies, alongside inquiries into AI consciousness and religious perspectives that highlight cultural and ethical considerations \cite{IsDeweys,alkhouri2024role,he2024artificial}.  
These diverse approaches underline the importance of contextualizing AI systems within broader humanistic and historical frameworks to assess their trustworthiness effectively. The integration of ethical, cultural, and philosophical dimensions in evaluating AI trustworthiness offers a more comprehensive understanding of its implications, enabling more informed and responsible development of AI systems.

\section{Methodology and Corpus Overview}
\subsection{Methodology}

We employ a Scoping Literature Review (SLR) following the guidelines by \citet{arksey2005scoping} to identify articles studying trustworthiness of AI systems. SLRs guide the gathering and identification of papers in a topic area for scrutiny \cite{kastner2012most}, which enables us to perform our thematic investigation of AIES \& FAccT scholarship. We use qualitative manual coding and computational corpus analysis, including topic modeling, to extract themes and patterns from the data.

\header{Data Collection} 
The data consists of peer-reviewed articles published by the AIES \& FAccT conferences between 2018–2025, downloaded from the ACM website on September 05, 2025\footnote{Note that the proceedings of AIES 2025 are not available when we have submitted this manuscript, hence AIES 2025 articles are excluded from this review.}. In total, 241 articles (83 AIES,  158 FAccT) are obtained that include the keyword ``trustworthiness'' in the full-text excluding references; 6 non-archival extended abstracts are excluded, leaving us 235 articles for screening.

\header{Screening and Selection Criteria}
We note several challenges in reviewing the AIES \& FAccT papers. The concept of trustworthiness is often embedded throughout papers instead of being the core theme. For example, auditing algorithmic systems or exploring potential biases in AI systems eventually help in understanding their trustworthiness, but this is often a point of discussion rather than the core contribution. Therefore, it is difficult to conclude that these articles engage with the core concept. Hence, we aim for a balanced approach studying trustworthiness, by including two types of article in our analysis: 
\begin{enumerate*}[label=(\roman*)]
\item articles that directly address trustworthiness as a central theme, like \citep{liao2022designing,Ferrario,jacovi,ferrario2025trustworthiness}, and 
\item articles where trustworthiness is not the main focus, but the authors explain how their findings contribute to a broader understanding of system trustworthiness and, ultimately, user trust. 
\end{enumerate*}

Therefore, keeping our balanced approach, following prior literature reviews on trust \citep{mehrotrareview,vereschak2021evaluate,benk2024twenty} and best research practices to study a particular topic \citep{ajmani,proferes2021studying}, we define our \emph{inclusion criteria} as:
\begin{enumerate}[label={(I\arabic*)},ref={I\arabic*}]
    \item One of the contributions of the article is linked to understanding trustworthiness of an algorithmic system.
    \item The article engages with trustworthiness as a component of their measure, directly influenced by the results, or as a key discussion theme.
\end{enumerate}

\noindent%
Our \emph{exclusion criteria} are:
\begin{enumerate}[label={(E\arabic*)},ref={E\arabic*}]
    \item The article discusses the need for AI trustworthiness without directly defining, measuring, or modeling it.
    \item The article is classified as a survey, scoping review, or literature review.
\end{enumerate}

\begin{figure*}[t!]
    \centering
    \includegraphics[width=0.5\linewidth]{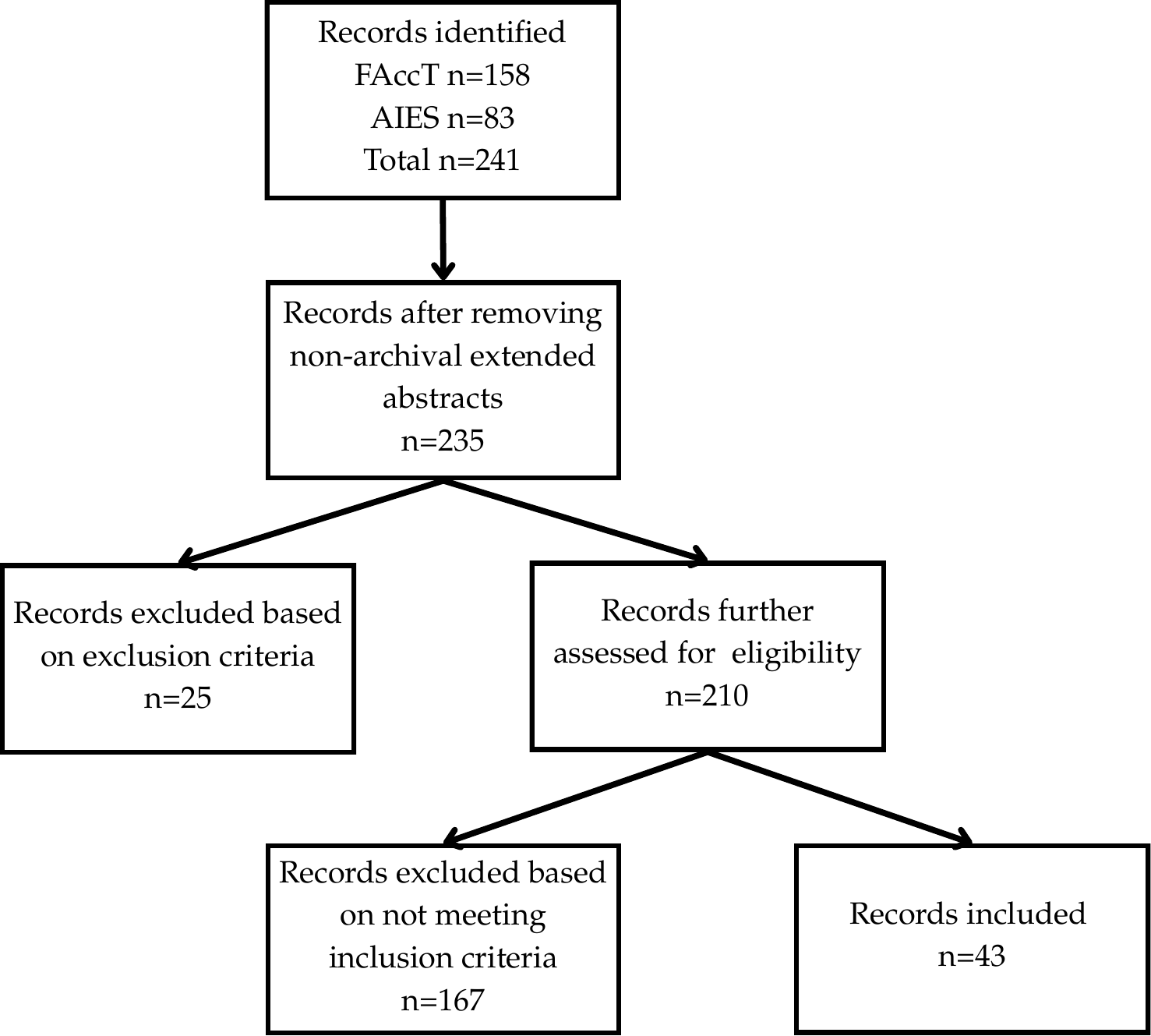}
    \caption{Flowchart of the articles reviewing process following the PRISMA protocol \cite{page2021prisma}.}
        \label{fig:mermaid}
\end{figure*}

\noindent%
The final corpus consists of 43 papers after applying our inclusion and exclusion criteria. An overview of our review process following the PRISMA protocol \cite{page2021prisma} is shown in Figure \ref{fig:mermaid}.

\header{Free-form Questions} 
Following the Goal-Question-Metric (GQM) approach \cite{caldiera1994goal}, we analyze the articles along the GQM dimensions that are phrased as questions:
\begin{enumerate*}[label=(\roman*),topsep=0pt]
\item what is the goal of the paper, 
\item what are different research questions related to this goal, and 
\item what are metrics to measure some aspects/factors of this goal? 
\end{enumerate*}
This framework provides us with a way to categorize the understanding of trustworthiness in the AIES \& FAccT  community in the form of generic free-form questions closely tied to the research questions from Section~\ref{sec:intro}.



We employ a thematic analysis to analyze the 43 articles and address our research questions. First, we develop a coding scheme to extract relevant information from each paper. For defining trustworthiness (\ref{rq:1}), we identify key terms, phrases, and concepts used across papers to formulate common definitional themes. 
We further link the identified themes with the International Organization for Standardization (ISO) standard 5723 on trustworthiness \cite{ISO_Trustworthiness_Vocabulary_2022}, allowing us to track frequency and depth of coverage. The ISO standard is chosen as a reference framework as it represents an internationally recognized benchmark for trustworthiness in technical systems~\citep{toney}. Furthermore, to understand the drivers of studying trustworthiness, we use both inductive and deductive coding approaches, first allowing themes to emerge naturally from the text then mapping these against common clusters. 

For trustworthiness measures (\ref{rq:2}), we develop a hierarchical coding structure to categorize assessment methods and validation \& verification techniques. For application scenarios (\ref{rq:3}), we employ open coding to identify emerging categories of use cases, followed by axial coding to establish relationships between different application contexts. 
Finally, to better understand the role of values in AI trustworthiness (\ref{rq:4}), we analyze intended, embodied, and realized values. We followed~\citet{van2020embedding}'s work to study the role of values. According to his work, intended values refer to the ethical principles and societal benefits that AI systems aim to achieve, embodied values are those integrated into their design and implementation, and realized values are the actual outcomes observed in practice. We use a two-stage coding process where we first identify explicitly stated intended values related to trustworthiness and then compare these against realized outcomes and embodied values. This involves creating pairs of intended-realized values and analyzing any gaps or alignments between them. 

Our coding process is iterative, with initial codes refined over multiple analysis rounds. To ensure reliability, three researchers independently code a subset of papers (25\%) and calculate an average Cohen’s Kappa coefficient ($\kappa = 0.62$). Through iterative discussions, they update their understanding, recode previous papers to include emergent concepts, include new papers in small batche of five and achieve improved reliability. The team collaborated for about two months to discuss findings and potential discrepancies in coding of the articles. Through multiple discussions, discrepancies were reevaluated and corrected, resulting in the final coefficient of $\kappa = 0.91$.

\subsection{Corpus Overview}

First, we perform a metadata analysis on the final corpus of 43 articles, focusing on understanding the trustor's background, the object of trustworthiness, variables, and study type.
Our meta analysis reveals that there are six interlinked trustors: users of a specific AI system (20),\footnote{The number in brackets denote the count of articles corresponding to a specific dimension.} 
citizens (12), developers (4), designers (3), practitioners~(2), and public administrators (2). As to the object of trustworthiness, the corpus revolves around studying trustworthiness of AI linked with data sources (19) and institutions developing the AI system (12). The role of trustworthiness as a study variable is almost equally divided among the papers as independent variable (21) and dependent variable (22). Finally, trustworthiness is studied normatively in only 1/3rd of the studies while the remaining 2/3rd studies it in empirical study design.

\begin{figure}[ht!]
    \centering
    \includegraphics[width=1\linewidth]{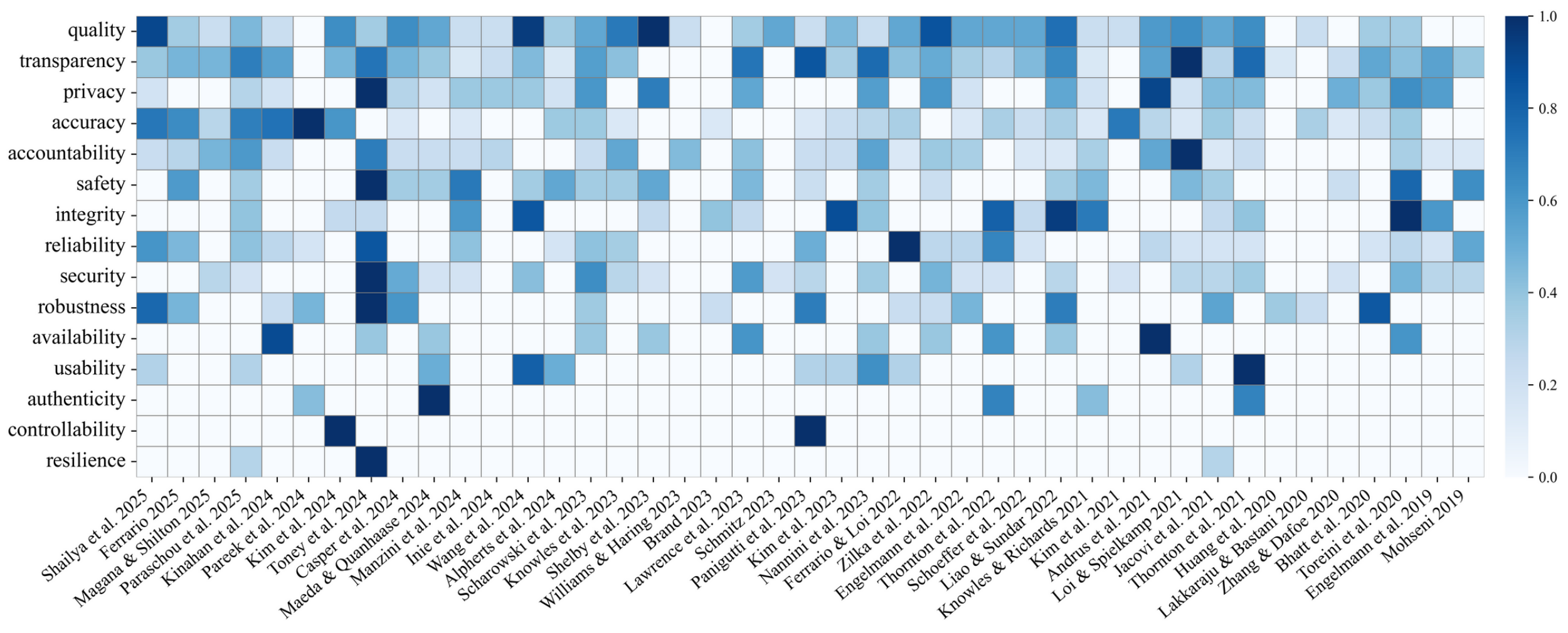}
    \caption{Heatmap of the normalized frequencies of trustworthiness dimensions in each paper of the final corpus ($N=43$).}
    \label{fig:tw_freq_heatmap}
\end{figure}

Second, Figure \ref{fig:tw_freq_heatmap} shows a heatmap of the log-normalized, min-max scaled word occurrences of trustworthiness dimensions by \citet{ISO_Trustworthiness_Vocabulary_2022}, in the full texts of the included corpus. Dimensions like quality, transparency, and accuracy occur frequently throughout the full texts, while authenticity, controllability, and resilience are mentioned relatively less. Some articles, like \citet{paraschou2025ties} and \citet{toney}, refer to a wide array of dimensions.

\begin{figure*}[t!]
    \noindent
    \begin{minipage}[t]{0.35\textwidth} 
        \vspace*{3pt} 
        \raggedright 
        Topic 1 (72):
        explanations, human, decision, language, study \\ \vspace{5pt}
        Topic 2 (85): transparency, accountability, public, human, technologies \\ \vspace{5pt}
        Topic 3 (49): fairness, bias, algorithmic, fair, discrimination \\  \vspace{5pt}
        Topic 4 (29): ethics, ethical, moral, responsible, norms
        \vspace*{\fill} 
    \end{minipage}%
    \hfill
    \begin{minipage}[t]{0.65\textwidth}
        \vspace{0pt} 
        \flushright
        \includegraphics[width=\linewidth, trim=0 0pt 0 0pt, clip]{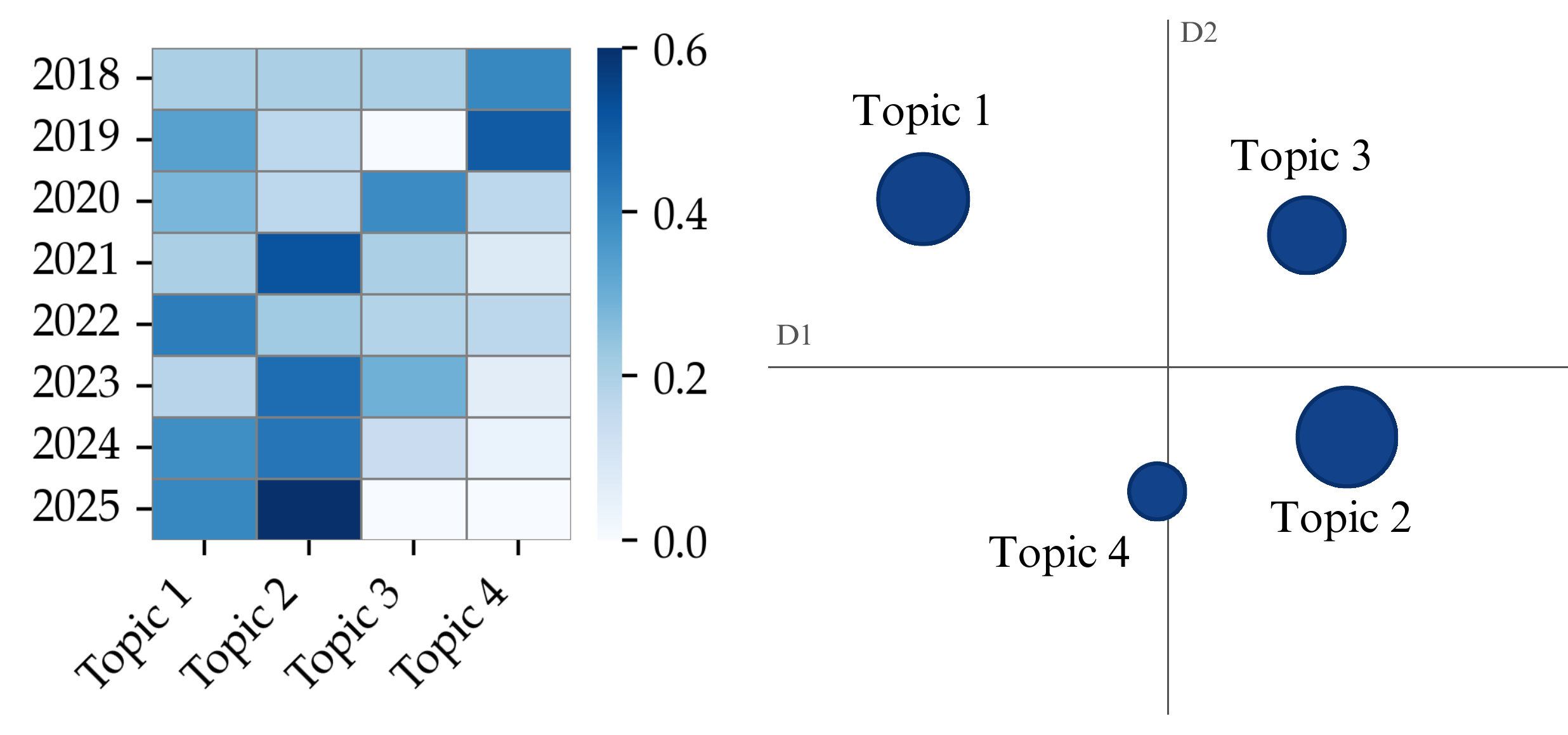}
    \label{fig:topic_model_full_vis}
    \end{minipage}
    \vspace{-17pt}
    \caption{Visualizations of topics generated using BERTopic applied to the initial keyword-based selection of articles ($N=235$). First, the heatmap (left) shows normalized topic distributions, grouped by year. Values refer to the proportion of documents associated with the topic. Second, the intertopic distance plot (right) shows the relative positions of topics in a 2D space.\protect\footnotemark } \vspace{-13pt}
    \label{fig:topic_model_vis}
\end{figure*}

Third, we broaden the analysis and discover themes across the initial keyword-based selection of 235 articles, without extended abstracts, through topic modeling. We apply BERTopic~\cite{grootendorst2022bertopic}, which leverages c-TF-IDF and the capabilities of the transformer-based language model BERT~\cite{devlin2018bert}, to analyze topics that are both coherent and contextually meaningful. The generated topics are analyzed using two visualizations as shown in Figure~\ref{fig:topic_model_vis}.
On the one hand, we present the normalized topic distributions over time using a heatmap to capture trends and shifts across publication years.
We observe a consistent prevalence and recent increase of topics relating to ``explanations'' and ``accountability'', while there is a relative decline in topics referring to ethics and norms.
On the other hand, we visualize the representations and relative distance of the topics in a 2D plot.
Notably, topics surrounding ``fairness,'' ``transparency,'' and ``ethics'' lie in closer proximity to each other, while they are more distanced from the topic around ``explanations.'' Also, we find that although fairness and bias are key ethical concerns, the considerable distance between fairness-related topics and ethical themes suggests that papers tend to emphasize one while giving limited attention to the other. 

\footnotetext{The representations of topics are reduced to two dimensions using the uniform manifold approximation and projection dimensionality reduction technique, enabling the generation of a 2D plot to visualize the topics.} 




\section{Understanding of AI Trustworthiness in the AIES \& FAccT  Community}


\subsection{Definition or Conceptualization}
\label{sec:definition}

To understand how to understand an AI system's trustworthiness, it is crucial to examine how we define or conceptualize it. \emph{Definition} refers to how trustworthiness is formally described and articulated in the selected papers and \emph{conceptualization} encompasses the broader understanding of trustworthiness, \eg considering its role in systems. 

First, the AIES \& FAccT community presents a rich and multifaceted understanding of AI trustworthiness, as evidenced across numerous studies. 
Through our coding and thematic analysis, we identify seven key conceptualization themes, summarized in Table~\ref{tab:definitions}. 

Analysis of the conceptual landscape reveals several critical insights about how AI trustworthiness is understood in contemporary scholarship. First, there exists a fundamental tension between transparency-focused approaches~(T1) and technical robustness perspectives (T6), suggesting competing philosophies about whether trustworthiness derives primarily from system interpretability or from demonstrated reliability regardless of internal opacity. Second, the prominence of anthropomorphization themes (T3) alongside ethical considerations (T4) indicates growing recognition that trust in AI cannot be divorced from human psychological tendencies to perceive social relationships with technology, a finding that challenges purely technical or procedural definitions. Third, the emphasis on broader societal implications (T5) and regulatory ecosystems signals an important shift from treating trustworthiness as an individual system property to understanding it as embedded within institutional contexts where harms, accountability, and public opinion are distributed unevenly across populations. Finally, the reliance on established frameworks like Mayer's model and NIST/ISO definitions (T7) alongside emerging concepts of perceived parasocial relationships suggests the field is simultaneously borrowing from organizational trust theory while grappling with AI-specific phenomena that existing models may not adequately capture. These insights collectively reveal that AI trustworthiness is conceptualized not as a unified construct but as a contested terrain where technical, psychological, ethical, and societal dimensions intersect and sometimes conflict.


\begin{center}
\captionsetup{type=table}
\vspace{-10pt}
\caption{Understanding of AI Trustworthiness: Key themes in definitions and conceptualizations.}
\label{tab:definitions}
\vspace*{-3mm}
\begin{tabularx}{\linewidth}{p{0.33\textwidth} X }
\toprule
 \textbf{Key themes} & \textbf{Sub-themes}  \\
\midrule
 T1: Emphasis on transparency as a foundational aspect of trustworthiness & Trustworthiness cues \citep{liao2022designing}, \textit{Good} explanation \citep{rayyan-64504518}, Open communication \citep{brandj}, Transparency, Provenance \& connections between them \citep{rayyan-64504562}, Computational reliabilism \& anti-monitoring \citep{Ferrario}, Expressing Uncertainty \citep{Sunnie2024a}, Transparency with human oversight \citep{rayyan-64504523}, Explanation Trustworthiness \citep{rayyan-64504630,shailya2025lext}, Transparency of algorithmic tools deployed in the criminal justice system \citep{rayyan-64504664} \\
 \midrule
T2: The importance of benchmarks, rigorous auditing and compliance mechanisms & White- and outside-the box audits \citep{Stephen2024}, Certification Labels \citep{nicolas}, Performance benchmarks and evaluation measures \citep{rayyan-64504620}, Mapping regulatory guidelines to philosophical account of accountability \citep{rayyan-64504624} \\
\midrule
T3: Perceived Anthropomorphization & (de)Anthropomorphized description of AI system \citep{Nanna2024}, Perceived Parasocial Relationships \citep{rayyan-64504506}, Facial AI inferences \citep{Severin2022}\\
\midrule
T4: Ethical considerations, including fairness and respect for user intent &  Inclusive \& intersectional algorithms \citep{kime}, Data leakage \& reproducibility \citep{Sean2024}, Model interpretability \citep{rayyan-64504617}, Understanding of harms and how
(and by whom) this has been informed \citep{Bran2023}, Interaction between risks and the dispositions of the trustee \citep{rayyan-64504660}, Informational fairness \citep{10.1145/3531146.3533218}, Achieving fairness \citep{rayyan-64504644}, Fairness by design \citep{rayyan-64504611}, Fair data procurement \citep{mckane}  \\
\midrule
T5: Broader societal implications: Regulatory Ecosystem and Accountability caused by harms & Social trustworthiness score \citep{Severin2019}, Trustworthiness criteria for policy-making \citep{rayyan-64504432}, Public trust in AI \& institutional trust \citep{Bran2021}, AI Assistant design \& Organizational practices and third-party governance \citep{rayyan-64504509}, sociotechnical harms \citep{rayyan-64504649}, Public opinion \citep{rayyan-64504662}, Safe, reliable, and acceptable to users and public \citep{magana2025frameworks}  \\
\midrule
T6: Justified reliance on AI’s reliability and technical robustness & Reasonable trust in model's output \citep{Umang2020}, Warranted Trust \citep{jacovi}, AI system must maintain to function correctly within its context of application \citep{ferrario2025trustworthiness}   \\
\midrule
T7: Mayer's \citep{mayer1995integrative} and Lee \& See's model \citep{lee2004trust} \& NIST/ISO definition & Ability, Benevolence \& Integrity \citep{Sunnie2023,liao2022designing,Sunnie2024a,wangr,lauren,rayyan-64504525,paraschou2025ties}, ABI+ (ABI, Predictability) framework \citep{toreini}, NIST \& ISO Definition \citep{toney} \\
\bottomrule
\end{tabularx} 
\addtocounter{table}{-1}
\end{center}


We also study the trustworthiness dimensions defined by the ISO trustworthiness standard \cite{ISO_Trustworthiness_Vocabulary_2022} to ensure that our analysis is based on internationally recognized standards.
`Transparency' is the most frequently mentioned ISO dimension, appearing in 21 out of 43
papers, followed by `accountability' and `explainability' in 12 papers. The less frequently mentioned ISO dimensions are `controllability' (5), `resilience' (4), `robustness' (4), `security' (4), `safety' (3), `reliability' (3), `usability' (3), and `provenance' (2). 

While many studies implicitly link trustworthiness to accuracy, reliability, transparency, and fairness, increasing attention is being given to more nuanced conceptualizations, such as stakeholder-centric approaches and systemic perspectives. Frequent references to ISO dimensions, particularly transparency, accountability, and explainability, underscore their importance in building trustworthy AI systems. However, a more comprehensive and standardized approach to defining and measuring AI trustworthiness is needed for the responsible development and deployment of AI systems.

\subsection{Drivers}
\label{sec:drivers}
\citet{anjomshoae2019explainable} have underlined the importance of considering the intended purpose when studying the trustworthiness of AI systems. To this end, this scoping review seeks to understand why the reviewed studies focus on understanding trustworthiness. Most of these works stated their motivation or intended purpose for studying trustworthiness. Table \ref{tab:DRIVERS} lists the drivers of the 43 papers included in the review, with some papers having more than one driver.

\bigskip

\begin{center}
\captionsetup{type=table}
\caption{The drivers of AI trustworthiness of the primary studies covered by the review.}
\label{tab:DRIVERS}

\vspace*{-3mm}
\begin{tabularx}{\linewidth}{p{0.2\textwidth} X}
\toprule
 \textbf{Drivers} & \textbf{Representative articles} \\
\midrule
D1: Conformity with regulations \& Guidelines and ethical considerations & \citep{Stephen2024,Severin2022,Sean2024,rayyan-64504518,rayyan-64504523,10.1145/3531146.3533218,nicolas,toney,brandj,rayyan-64504611,rayyan-64504620,rayyan-64504624,rayyan-64504644,rayyan-64504649,rayyan-64504562,rayyan-64504434,Bran2021,rayyan-64504630,rayyan-64504611,rayyan-64504664,toney,mckane,shailya2025lext,ferrario2025trustworthiness} \\
\midrule
D2: Promoting honesty & \citep{Severin2019,rayyan-64504660,rayyan-64504432} \\
\midrule
D3: Trust & \citep{Umang2020,Ferrario,Nanna2024,jacovi,Sunnie2024a,Sunnie2023,liao2022designing,rayyan-64504506,rayyan-64504509,rayyan-64504562,10.1145/3531146.3533218,rayyan-64504525,lauren,wangr,rayyan-64504617,rayyan-64504630,Bran2023,jacovi,rayyan-64504660,rayyan-64504432,Bran2021,toreini,rayyan-64504662,magana2025frameworks,paraschou2025ties} \\
\midrule
D4: Economic growth \& Cultural values & \citep{Severin2019,brandj} \\
\midrule
D5: Reproducibility  & \citep{Sean2024} \\
\midrule
D6: Accuracy & \citep{Aida2023,kime,rayyan-64504432} \\
\bottomrule
\end{tabularx} 
\end{center}

\addtocounter{table}{-1}

Increasing users’ trust in the system, ethical considerations, and conformity with regulations such as those of~\cite{EU_AI_Ethics_Guidelines, Canada_AI_Ethics_Principles, Canada_Directive_on_Automated_Decision_Making, Canada_Responsible_Use_of_Artificial_Intelligence_Guiding_Principles, Japan_Governance_Guidelines_for_Implementation_of_AI_Principles, UK_ICO_Guide_on_AI_and_Data_Protection, US_Presidential_Executive_Order_13960, US_White_House_OMB_Guidance_for_Regulation_of_Artificial_Intelligence_Applications}, 
and accuracy are among the listed motivations for the explanations. The table reveals that trust and ethical considerations are the most prominent drivers of studying AI trustworthiness. Naturally, trust and trustworthiness go hand in hand to increase users’ confidence in the system by understanding how its reasoning mechanism works \cite{anjomshoae2019explainable}. In applications requiring human-AI interaction, honesty and accuracy are among the main drivers for trustworthiness to ensure that AI’s decision-making is reliable and fair \cite{10.1145/3531146.3533218,rayyan-64504525}. For public AI systems, trustworthiness drivers are often linked with conformity with regulations and guidelines, and economic growth and cultural values \cite{Severin2019,brandj,Bran2021,rayyan-64504624,rayyan-64504662}. Finally, \citet{Sean2024} identify reproducibility of their results as a driver for studying trustworthiness for their system. Overall, these studies provide a holistic overview of the key drivers for studying AI trustworthiness.


\subsection{Measurement and Verification \& Validation}
\label{sec:measure}
\citet{jacobs2021measurement} have introduce measurement theory for AI/ML systems, where validation and verification are key to ensure we are measuring what we are trying to measure.
In the context of AI trustworthiness, measurement refers to quantifying specific attributes or characteristics of the AI system, typically aligned with ISO trustworthiness dimensions.
Following ISO 9001~\cite{dnv2015iso}, verification involves confirming through examination and provision of objective evidence that the specified trustworthiness dimensions are fulfilled.
Validation is similar to verification, but must be confirmed under real-world usage conditions.

\header{Measurement}
To analyze the measurement techniques for AI trustworthiness, we follow \citet{schlicker2025we}'s distinction of actual and perceived trustworthiness. \citet{schlicker2025we} define a system’s actual trustworthiness as a latent construct that indicates the true value of a system’s trustworthiness (in alignment with Realistic Accuracy Model \cite{funder1995accuracy}), \eg benevolence, integrity, and ability \cite{mehrotrareview,lee2004trust,mayer1995integrative}. Similarly, perceived trustworthiness reflects the result of a trustor’s assessment of the trustee’s actual trustworthiness. 

\begin{table*}[t!]
\caption{The measurement of AI trustworthiness of the primary studies covered by the review.}
\label{tab:Measurement}
\centering
\begin{tabular}{p{0.3\textwidth} p{0.65\textwidth}}
\toprule
 \textbf{Measurement} & \textbf{Type of measure} \\
\midrule
 Likert scale & McKnight et al.[\citet{HARRISONMCKNIGHT2002297}] \citep{Sunnie2024a}, Jian et al. [\citet{Jian01032000}] \citep{rayyan-64504525},  Perceived trustworthiness (\textit{Own scale}) \citep{10.1145/3531146.3533218} \\
\midrule
Reliance behavior & Change in people’s trust judgment \citep{liao2022designing}, Agreement between a participant’s answer and that of the AI system \citep{Sunnie2024a,Sunnie2023,magana2025frameworks} \citep{rayyan-64504525,wangr,rayyan-64504620} \\
\midrule
Actual cues (\eg precision, recall, and f1-scores) & Positive predictive value \cite{kime}, LDA topic modelling \citep{Severin2019}, Stochastic model with surrogates \citep{rayyan-64504518}, \citep{Aida2023,toreini,Sean2024,Stephen2024,rayyan-64504523}, Metrics for plausibility and faithfulness scores \citep{shailya2025lext} \\
\midrule
Assessment of impact and alignment with values and concerns & Ethical charters \& internal review bodies \citep{rayyan-64504509}, sociotechnical harms taxonomy, Moral groundings for public transparency \citep{rayyan-64504624} \citep{rayyan-64504649} \\
\midrule
 Levels of monitoring activities & Monitoring-avoiding relation between trustor \& trustee \citep{Ferrario} \citep{toreini}, Trustworthiness level across system accuracy ranges \citep{ferrario2025trustworthiness}\\
 \midrule
Communication style and anthropomorphic features & Roleplaying \& reciprocal engagement \citep{rayyan-64504506} \\
 \midrule
Qualitative assessment & Interviews \& Surveys  \citep{nicolas,Umang2020,rayyan-64504562,Nanna2024,rayyan-64504617,wangr,rayyan-64504432,rayyan-64504662,Severin2022,mckane}, User satisfaction \citep{rayyan-64504664}\\
\midrule
Model Cards and AI Assessment Catalog & Information about data leakage \& selective inference for EEG \cite{Sean2024}, AI Assessment
Catalog \citep{rayyan-64504644}, NIST AI Risk Management Framework \& other national frameworks \citep{toney} \\
\midrule
Mediator \& Trustee role & Social, Institutional and Temporal Embeddedness, Situational Normality, Structural Assurance, Motivation and ABI \cite{lauren}\\
\bottomrule
\end{tabular} \vspace{-10pt}
\end{table*}

Analysis of measurement strategies reveals several important insights about the operationalization of AI trustworthiness. First, the field exhibits a problematic conflation between measuring trust (a psychological state) and measuring trustworthiness (system properties that warrant trust). 
Reliance behavior and Likert scales capture user perceptions while technical metrics assess objective capabilities, yet the relationship between these remains theoretically underdeveloped. 
Second, the diversity of approaches, from LDA topic modeling to ethical charters to role-playing exercises, suggests that trustworthiness is inherently multi-faceted and resists reduction to single metrics, yet this proliferation also indicates a lack of consensus on what actually constitutes valid measurement. 
Third, the emergence of documentation artifacts like model cards~\cite{Sean2024} and AI assessment catalogs~\citep{rayyan-64504644} represents an important shift toward longitudinal, context-aware evaluation rather than snapshot assessments, implicitly acknowledging that trustworthiness is not static but emerges through ongoing scrutiny. 
Finally, the inclusion of mediator and trustee role factors (social embeddedness, situational normality, structural assurance) indicates growing sophistication in recognizing that measurement must account for the relational and contextual dimensions of trust rather than treating it as residing solely in the AI system. Together, these developments suggest the field is moving toward more holistic measurement paradigms, though integration across approaches remains elusive.

\header{Verification and/or Validation}
More than half (34 out of 43) of the papers did not (explicitly) provide or adopt verification or validation approaches.
Among the remaining 9 papers, surveys are the most widely used verification methodology, appearing in 5 papers~\citep{Severin2022,Nanna2024,10.1145/3531146.3533218,rayyan-64504525,rayyan-64504662}.
E.g., \citet{rayyan-64504525} conduct a survey-based between-subject experiments involving 300 participants to investigate trust development, erosion, and recovery during AI-assisted decision-making.
Additionally, \citet{rayyan-64504662} pre-register their experiments to ensure methodological rigor and transparency.
\citet{rayyan-64504617} and \citet{rayyan-64504432} verify trustworthiness measurement through user studies with domain experts.
Finally, only one paper employs a mixed method consisting of qualitative interviews and a subsequent survey to collect quantitative data~\citep{nicolas}.


The gap in current research, \ie the insufficient attention to validation under real-world constraints and the lack of alignment between verification practice and trustworthiness dimensions, can be attributed to two primary factors: the lack of real-world data or representative scenarios~\citep{rayyan-64504434} 
and the absence of multi-disciplinary collaboration 
~\citep{brundage2020toward}. The 9 papers that do so represent important but limited work in controlled settings, not diverse real-world environments. Our thoughts resonate with \citet{rayyan-64504434} and \citet{brandj} who have demonstrated challenges in obtaining representative datasets and how siloed approaches can miss critical sociotechnical dimensions for verification or validation. 

\subsection{Application Scenarios}
\label{sec:application}
AI techniques are applied across domains, including high-stakes ones, making trustworthiness vital to their success.
The AIES \& FAccT community demonstrates a comprehensive and diverse exploration of trustworthiness across various application scenarios; see Table~\ref{tab:Scenarios}.
High-stake domains, including healthcare and medical applications, financial and economic systems, human resources, governance, law, and justice, are the most prominent application scenarios of studying trustworthiness of AI systems (22 out of 43).
12 out of 43 papers do not centre on a specific application scenario but rather generally discuss AI applications and their impact on society. 
Additionally,  \citet{rayyan-64504518} and \citet{rayyan-64504523} examine AI policies with a focus on explainability of AI systems.
\citet{rayyan-64504518} perform a thematic and gap analysis of policies and standards on explainability in EU, US, and UK and provide a set of recommendations on how to address explainability in regulations for AI systems.
\citet{rayyan-64504523} analyze the EU AI Act~\citep{european2021proposal} and argue that it neither mandates explainable AI nor bans the use of black-box AI systems; instead, it seeks to achieve its stated policy objectives with a focus on transparency and human oversight.

\bigskip

\begin{center}
\captionsetup{type=table}
\caption{The application scenarios of AI trustworthiness of the primary studies covered by the review.}
\label{tab:Scenarios}
\vspace*{-3mm}
\begin{tabularx}{\linewidth}{p{0.25\textwidth} X}
\toprule
 \textbf{Application Scenarios} & \textbf{Representative articles} \\
 \midrule
 Healthcare and medical applications & \citep{liao2022designing,nicolas,toreini,Sunnie2024a,Sean2024,Vivian2023,rayyan-64504434,mckane}\\ 
 \midrule
 Financial and economic systems & \citep{Umang2020,brandj,rayyan-64504434,10.1145/3531146.3533218,nicolas,mckane} \\
 \midrule
 Human resource & \citep{Severin2022,nicolas,rayyan-64504630,rayyan-64504434,mckane} \\
 \midrule
 Governance, law, and justice & \citep{rayyan-64504620,rayyan-64504624,rayyan-64504630,rayyan-64504662,rayyan-64504611,rayyan-64504617,Severin2019,rayyan-64504664,rayyan-64504660,rayyan-64504644,Bran2021} \\
 \midrule
 Content moderation and information systems & \citep{Umang2020,Chiara2022}\\
 \midrule
 Personal assistance and interaction & \citep{Lydia2024,rayyan-64504509,nicolas}\\
 \midrule
 Environmental science & \citep{rayyan-64504432,Sunnie2023,rayyan-64504562} \\
 \midrule
 Domain agnostic & \citep{Stephen2024,Nanna2024,jacovi,Bran2023,Bran2021,rayyan-64504525,toney,kime,rayyan-64504649,rayyan-64504506,lauren,toreini,rayyan-64504630,Severin2022} \\
 \midrule
  AI policies and standards & \citep{rayyan-64504518,rayyan-64504523} \\ 
  \midrule
  Facial recognition & \citep{kime} \\ 
  \midrule
  Software engineering and programming & \citep{wangr} \\ 
 \bottomrule
\end{tabularx}
\addtocounter{table}{-1}
\end{center}

\subsection{Intended, Embedded, and Realized Values}
\label{sec:values}
Our analysis reveals a significant gap between the intended, embodied, and realized values of AI systems. While many are developed with good intentions, \eg promoting transparency, fairness, and accountability, these values are not always reflected in the final product or its impact on society.

\header{Intended Values}
The most frequently cited intended values were transparency (26), fairness (19), and accountability (12)\footnote{39 articles had more than one intended value for studying AI trustworthiness.}. Other important values included privacy (8), security (5), and human oversight (5). E.g., \citet{Umang2020} emphasize the importance of transparency and trustworthiness as the intended values of explainable AI. Similarly, \citet{Stephen2024} highlight the role of AI audits in identifying risks and improving transparency, and \citet{Severin2019} examine the intended value of the social credit system to promote honesty and trust in Chinese society. In abidance with the AI act, \citet{rayyan-64504523} examine the role of explainable AI and state the intended value to ensure AI is trustworthy by focusing on transparency and human oversight. This includes enabling users to interpret outputs and manage risks associated with AI systems.

\header{Embodied Values}
The embodied values in studying AI trustworthiness are often shaped by the techniques used, the stakeholders involved, and the context of deployment. \citet{rayyan-64504518} examine the embodied value of explainable AI, which is determined by how well it can fulfill its intended purpose without compromising other desiderata, such as accuracy and privacy. \citet{nicolas} examine the embodied value of certification labels, which is determined by how well they can fulfill their intended purpose by reflecting the AI system's compliance with trustworthiness criteria. \citet{rayyan-64504432} examine the embodied value of AI models, determined by how well they can generate explanations that help understand the relations between visual elements in street view imagery and socio-economic variables such as housing pricing.

\header{Realized Values}
The realized values of AI systems can differ significantly from their intended values due to various factors, including design flaws, unintended consequences, and societal biases. Some examples of how realized values can deviate from intended values: \citet{Ferrario} find that the realized value of explainability in AI systems may differ from its intended value, particularly when explainability does not lead to a reduction in monitoring or when it fosters an unjustified belief in the AI's trustworthiness and \citet{Sunnie2023} find that the realized value of AI systems may differ from their intended value due to factors such as malicious manipulation of trustworthiness cues or ill-designed cues that mislead users.

The gap between intended, embodied, and realized values in AI systems raises important questions about the effectiveness of current approaches to embedding values in AI design for studying AI trustworthiness. 
It highlights the need for more robust methods to evaluate AI's societal impact.
\section{Discussion: Challenges, Open Questions, and Future Directions}
\label{sec:discussion}

This review has examined the multifaceted nature of AI trustworthiness as articulated within the AIES \& FAccT community. This section examines the challenges, open questions, and future directions that arise from our analysis.

\subsection{Why Do We Need to Rethink AI Trustworthiness?}
Several critical gaps and limitations emerge from our analysis in Section~\ref{sec:definition}. Despite this rich conceptual terrain, the current literature inadequately address: who defines trustworthiness criteria, whose harms count as significant, and how user intent is interpreted remain largely unexamined despite ethical considerations being prominent. Additionally, while there is convergence around components like accuracy and fairness linking to ISO dimensions, the literature largely neglects the power dynamics and structural inequalities that fundamentally shape trust relationships in AI systems \cite{dobbe2021hard}.

Additionally, while benchmarks and auditing mechanisms (T2) are emphasized, there is insufficient attention to how compliance-focused approaches might incentivize ``ethics washing'' where organizations meet procedural requirements without substantively addressing underlying trust deficits. The conceptualizations also struggle with context-dependency: trustworthiness defined through computational reliability may be necessary but insufficient in domains like criminal justice (as noted in T1) where procedural justice and legitimacy matter as much as accuracy, yet the frameworks provide limited guidance on how to weight these competing demands situationally.

Current conceptualizations, despite their evolution from purely technical metrics to sociotechnical constructs, fail to fully account for how trust can erode or strengthen over time, particularly in response to system failures or successful interventions \cite{dzhelyova2012temporal}. An emphasis on system-level characteristics and user perceptions, while important, has come at the expense of examining broader institutional and systemic factors that influence trustworthiness, such as regulatory frameworks, corporate governance structures, and market incentives. 
Without addressing these gaps, current approaches to building trustworthy AI risk perpetuating existing biases and power imbalances while failing to establish sustainable trust relationships across user communities~\cite{osasona2024reviewing}. 

Based on our analysis of the corpus, we propose the following measures:
\begin{enumerate}[label=(\arabic*)]
    \item Future research must implement longitudinal studies that track trust dynamics over time, and design systems with mechanisms to rebuild trust after failures rather than focusing solely on initial trust establishment.
    \item At the heart of AI trustworthiness lies a fundamental distinction between perceived and actual trustworthiness \cite{schlicker2022calibrated}. Future research should clearly distinguish them to foster a more unified understanding of AI trustworthiness.
    \item Develop comprehensive trustworthiness frameworks that explicitly incorporate institutional accountability measures, regulatory compliance mechanisms, and transparent corporate governance structures.
\end{enumerate}
\subsection{Is AI Trustworthiness Just a Checklist?}
A critical analysis of the drivers of AI trustworthiness revealed oversights in how the field currently conceptualizes motivations for studying this crucial aspect. While the reviewed literature identifies several key drivers (see Section~\ref{sec:drivers}), the understanding appears superficial and fails to address several fundamental concerns.

First, the heavy emphasis on regulatory compliance and ethical guidelines, while important, suggests a potentially reactive rather than proactive approach to trustworthiness \cite{tyler2001trust}. Many studies treat regulations as a checklist (see representative articles for D3 in Table~\ref{tab:DRIVERS})
rather than engaging with the deeper philosophical and societal implications of AI trustworthiness. This compliance-driven approach risks creating a false sense of security while potentially missing emerging challenges not yet addressed by current regulations, such as the evolving nature of biases in dynamic systems, the challenge of ensuring explainability in complex models \cite{saeed2023explainable}, attributing accountability in autonomous decision-making \cite{busuioc2021accountable}, and impacts on human agency and societal power structures \cite{santoni}. In real-world industry settings, compliance requirements often act as crucial entry points and motivators for responsible AI work \cite{diaz2023connecting}. When organizational resources and attention are limited, framing trustworthiness as a compliance issue can secure support and budget allocations \cite{caldwell2003organizational}. 

Second, the classification of drivers reveals a bias toward accuracy and promotion of trust in AI, with a limited focus on end-user needs and societal impacts. While ``trust'' is frequently cited as a driver, there is an inadequate exploration of building an appropriate level of trust in AI, resulting in avoidance of over- and under-trust~\cite{10.1145/3610578,mehrotrareview}. 

Third, we found vagueness in debates about the safety behavior of AI systems and identifying \& diagnosing safety risks in complex social contexts. Safety has been designated as a key component of AI trustworthiness both in the EU AI Act \cite{european2021proposal} and ISO \cite{ISO_Trustworthiness_Vocabulary_2022} \& NIST \cite{NIST_AI_Risk_Management_Framework} guidelines. However, in our analysis, system safety does \emph{not} appear as a prominent measure for trustworthiness except for the mathematical formalism \cite{toney}. This raises two key questions: (i)~Is safety marred by an underlying vagueness where it is hard to establish whether a system is safe or not \cite{dobbe2021hard}? And (ii) does it even make sense to study trustworthiness of an AI system without exploring its safety especially from a sociotechnical lens?

Finally, drivers related to long-term sustainability, environmental impact, and social justice are limited in our analyzed papers. While accuracy is listed as a driver, there is limited attention to the complex relationship between technical accuracy and real-world effectiveness. This limited understanding of drivers profoundly impacts how trustworthiness is studied and implemented in AI systems. Without a more nuanced understanding of why trustworthiness matters, the field risks producing solutions that address surface concerns while failing to engage with deeper systemic challenges. Hence, we propose the following measures:

\begin{enumerate}[label=(\arabic*)]
    \item The ideal approach could involve using compliance frameworks as practical starting points while simultaneously cultivating organizational cultures that recognize the broader societal implications of AI systems beyond regulatory requirements. 
    \item Develop frameworks and metrics that address the appropriateness of trust, ensuring users understand both the capabilities and limitations of AI systems they interact with.
    \item Future research must expand beyond these conventional drivers of trustworthiness to include  societal concerns, power dynamics, and long-term implications for human-AI interaction.
\end{enumerate}

\subsection{Measurement Problem in AI Trustworthiness} 
The critical analysis of measurement, verification, and validation approaches in AI trustworthiness research in Section~\ref{sec:measure} reveals gaps and methodological shortcomings to establish robust frameworks for evaluating trustworthiness of AI systems and build appropriate level of trust. More precisely, this necessitates: 
\begin{enumerate*}[label=(\roman*)]
\item conducting evaluations of both actual and perceived trustworthiness, \ie establishing valid and reliable measurements, 
\item identifying methods to make valid comparisons between these two dimensions, and also 
\item to achieve alignment between them. 
\end{enumerate*}

First, current research presents a significant gap concerning evaluation of trustworthiness. Prior studies have described the general challenge of aligning actual and perceived trustworthiness \cite{mehrotrareview, liao2022designing}. They have also identified relevant components that constitute actual and perceived  trustworthiness \cite{schlicker_trustworthy_2025}. Additionally, previous work has emphasized the necessity for a communication process that conveys information about an AI system and its actual trustworthiness to shape and align trustworthiness perceptions \cite{liao2022designing}. However, we still lack understanding of how the evaluation of actual trustworthiness and the successful alignment of this actual trustworthiness with perceived trustworthiness would manifest in practice, based on concrete measures and quantitative values for trustworthiness characteristics both for the AI system and user perceptions.

Second, when trustworthiness is measured using only questionnaires or reliance behaviour, there is still uncertainty about the trustee’s actual trustworthiness \cite{schlicker2025we}. Robust methodologies that incorporate objective measures of trustworthiness alongside subjective assessments are essential for developing a comprehensive understanding of trustworthiness in human-AI interaction. However, it is still complex to consider that actual and perceived trustworthiness combined together provides a clear overview of the AI system's overall trustworthiness, perhaps the $2+2=4$ analogy is not a fit here. We believe the issue is there is no single correct way of doing this, \textit{i.e.}, making individual characteristics commensurable. There is no ground truth to questions like what is the objective trustworthiness value for a given performance or explainability score of an AI system \cite{Aida2023}.  

The third problem differs from the first two in a fundamental way. Rather than dealing with how to combine or compare trustworthiness measures and individual traits, it addresses a more basic question: which specific characteristics should be evaluated when assessing trustworthiness in the first place? The challenge here isn't about creating compatible measurement scales. Instead, it involves establishing comparable sets of characteristics between two sides—the AI system itself and the users who perceive it. Critical questions arise from this challenge: Which performance metrics should be considered? What aspects of transparency matter? How should explainability be evaluated? What privacy measures and safety standards should factor into the assessment of AI system trustworthiness? These questions are all fundamentally connected to this obstacle.

The verification and validation issues are even more concerning, with nearly half of the papers failing to address these crucial aspects. For example, what is the source of cues that inform users about an AI system's trustworthiness and how can we verify or validate it?
One perspective holds that these cues should originate directly from the systems and their creators. Some researchers advocate for defining good cues that technology developers should implement \cite{liao2022designing}. These same scholars acknowledge that such indicators must be accurate and honest to be effective. When examining practical applications, they assume this honesty requirement is already met by creators and focus their analysis on additional necessary conditions.
This assumption proves problematic in real-world scenarios. System providers cannot simply be expected to present entirely accurate cues about their products' limitations. Commercial entities operate with sales objectives as their primary driver. When users place excessive confidence in a product beyond what's justified, achieving proper confidence calibration would require the provider to communicate information that reduces trust. Yet commercial logic creates a strong disincentive for broadcasting unfavorable information; doing so would directly harm revenue and competitive positioning. It seems unlikely, therefore, that providers would intentionally engineer their systems to broadcast cues indicating mediocre or poor trustworthiness to potential users. 

We propose the following measures for effective measurement, validation and verification:

\begin{enumerate}[label=(\arabic*)]
    \item Our analysis suggests a pressing need for AI system designers and developers to must explain their rationale for choosing particular trustworthiness attributes and specific metrics to measure those attributes. They also need to justify their methodology for prioritizing and combining these elements to generate a composite trustworthiness score. In doing so, they should be required to acknowledge their inherent perspectives and the criteria they apply when judging an AI system's trustworthiness.
    \item Greater focus on trustworthiness assessment mechanisms within the trust development process would likely enhance future research efforts to clarify how perceived trustworthiness transforms into actual trust and subsequent trusting behaviors.
    \item  Those who provide the cues to the AI system users need to elaborate on and give reasons why certain characteristics of the AI system and trustworthiness cues warrant trust. Also, focus on understanding how cues are relevant for the users, how they detect and utlize it forms a key element for verification and validation of the trustworthiness assessment.
\end{enumerate} 


\subsection{Values Gap in AI Trustworthiness}
Our critical examination of the values in AI trustworthiness research reveals significant oversights and problematic assumptions that limit our understanding of how values materialize in practice. The analysis of intended values demonstrates a concerning preoccupation with surface-level technical attributes such as transparency and fairness, while giving insufficient attention to deeper systemic values such as social justice, environmental sustainability, and cultural preservation. This narrow focus reflects a persistent bias toward engineering solutions rather than addressing fundamental societal challenges \cite{olteanu2019social}. The heavy emphasis on transparency
risks becoming a performative gesture rather than a meaningful commitment to  openness and accountability, thereby risking to entrench problematic capture of the institutions that are meant to support public values \cite{10.1145/3531146.3533241,10.1145/3488666}.

The discussion of embodied values highlights weaknesses in current research. The literature often oversimplifies how values are embedded in technical systems, reducing complex social and ethical considerations to fit technical implementations. 
This simplification risks a false sense of progress while neglecting the deeper challenges of value implementation \cite{stern2002eva}. Particularly concerning is the treatment of realized values, where the analysis lacks a systematic framework for understanding value divergence. The literature notes discrepancies between intended and realized values but inadequately theorizes these gaps, attributing them to superficial issues like ``design flaws'' or ``unintended consequences'' without addressing structural barriers that hinder the realization of intended values.

The conclusion that a ``more holistic approach'' is required appears insufficient given the complexity of the challenges at hand. 
We need rigorous, dynamic frameworks reflecting the contested and evolving nature of values in AI systems.
Such frameworks must move beyond static, universal notions of values to address how they are negotiated, implemented, and evaluated in real-world contexts, confronting structures of power. E.g., these power structures manifest in multiple dimensions that trustworthiness research must address: 
\begin{enumerate*}[label=(\roman*)]
    \item corporate concentrations of AI development capacity that privilege certain stakeholders' definitions of ``trustworthiness,''
    \item institutional hierarchies within organizations that determine whose values and concerns shape AI systems,
    \item socio-economic disparities that affect who benefits from or bears risks of AI deployment, and
    \item geopolitical imbalances that enable dominant nations to set global AI governance norms; and professional authority structures that privilege technical expertise over lived experience.
\end{enumerate*} 

We recommend the following measures to reduce the values gap in AI trustworthiness:
\begin{enumerate}[label=(\arabic*)]
    \item Implementing participatory design methodologies that meaningfully involve marginalized communities in defining trustworthiness criteria as informed by \citet{harrington2019deconstructing}.
    \item Developing evaluation frameworks that assess differential impacts across diverse populations rather than assuming universal benefits \cite{carey2017glossary}.
    \item Establishing transparency mechanisms that reveal how power influences AI system development decisions and provide mechanisms to contest them as showcased by \citet{ehsan2021expanding,umap}.
    \item Creating accountability structures that redistribute decision-making authority beyond 
    technical experts \cite{busuioc2021accountable} and recognizing that trustworthiness is a contested concept shaped by those with the power to define it \cite{hardin2002trust}.
\end{enumerate}

\subsection{Operationalizing Recommendations: Use Cases and Transformations}

In the above four subsections, we propose general recommendations and measures for future directions from a sociotechnical perspective. Concretely, these recommendations can be deployed across multiple AI application domains. To make them more actionable for researchers working from technical perspectives, such as machine learning practitioners, we provide several illustrative examples across diverse applications: AI-powered clinical decision support systems and conversational agents for caregiver mental well-being support (AI in healthcare), AI-powered hiring/recruitment systems (AI in recruitment), criminal risk assessment algorithms (AI in criminal justice), and AI-powered adaptive learning platforms and automated assessment systems (AI in education). Table~\ref{tab:use_cases} demonstrates how current AI systems are defined, built, and evaluated in these use cases, and illustrates the transformative changes that our proposed recommendations and measures would bring to each domain.


\bigskip

\begin{center}
\captionsetup{type=table}
\caption{Comparison of current AI system practices and practices informed by our proposed recommendations: Illustrative use cases shows how our recommendations address limitations in current practices.}
\label{tab:use_cases}
\small
\vspace*{-3mm}
\begin{tabularx}{\textwidth}{p{0.14\textwidth}  p{0.25\textwidth} X}
    \toprule
        \textbf{Applications} & \textbf{Current systems} & \textbf{What our recommendations bring} \\
        \midrule
        AI-powered clinical decision support system & Evaluate overall accuracy metrics (sensitivity/specificity) on aggregated test datasets~\citep{hicks2022evaluation}; Measure trust through post-deployment user surveys with general practitioners~\citep{chen2022acceptance}; Apply fairness metrics (e.g., demographic parity, equalized odds) to measure performance disparities across protected attributes like race and gender~\citep{chen2023algorithmic}; Assess system performance using standardized benchmark datasets primarily from well-resourced healthcare settings~\citep{blagec2023benchmark}. & \textbf{Participatory measurement design}: Co-develop trustworthiness criteria with patients from underserved communities (e.g., rural populations, minorities with higher disease prevalence) to identify what "trustworthy screening" means beyond accuracy—including accessibility, cultural sensitivity, and explanation comprehensibility; \textbf{Differential impact evaluation}: Stratify performance metrics across race, socioeconomic status, age, and geographic location to reveal disparities (e.g., higher false negative rates in darker skin tones due to training data bias); \textbf{Power-aware transparency}: Document and disclose how decisions about training data selection, feature prioritization, and deployment locations were made, revealing potential bias toward well-resourced hospitals; \textbf{Redistributed evaluation authority}: Include community health workers, patients, and advocacy groups in defining evaluation success criteria alongside ophthalmologists and data scientists; \textbf{Context-sensitive trustworthiness}: Recognize that trustworthiness criteria differ—rural clinic practitioners may prioritize offline functionality and low-bandwidth operation, while medical centers focus on integration with their existing systems. \\
        \midrule
        Conversational agents for caregiver mental well-being support & Maintain system safety mainly through content moderation filters~\citep{sarkar2023review}; Measure user engagement metrics (session duration, return rate, sentiment scores)~\citep{limpanopparat2024user}; Assess conversational quality using expert annotations and standardized mental health screening tools~\citep{dosovitsky2021psychometric}; Conduct one-time user satisfaction surveys post-interaction~\citep{limpanopparat2024user}; Focus on clinical efficacy benchmarks and harm prevention protocols~\citep{sarkar2023review}. & \textbf{Longitudinal trust dynamics}: Track how caregiver trust evolves across crisis moments (e.g., after receiving unhelpful advice during acute stress) and design explicit trust repair mechanisms, such as human handoff protocols when the system fails to provide adequate support; \textbf{Perceived vs. actual trustworthiness}: Distinguish between caregivers' confidence in the system (perceived) and its actual clinical appropriateness—measuring whether users over-rely on the agent; \textbf{Institutional accountability frameworks}: Establish clear governance structures defining who is responsible when the system provides harmful advice—including regulatory compliance mechanisms, clinical oversight boards, and transparent corporate policies on data usage and care escalation; \textbf{Appropriateness of trust calibration}: Develop frameworks that help caregivers understand when to trust the AI (e.g., emotional validation, stress management techniques) versus when to seek human professionals (e.g., suicidal ideation, severe depression), with explicit system boundaries \\
        \midrule
        AI-powered hiring/recruitment systems & Evaluate prediction accuracy by measuring how well AI selections match historical hiring decisions and subsequent employee performance~\citep{black2020ai}; 
        Assess system efficiency in filtering large applicant pools~\cite{chen2023collaboration}; Apply minimal fairness audits primarily to meet legal compliance requirements~\citep{equal2023assessing}. 
        & \textbf{Participatory measurement design}: Co-develop trustworthiness criteria with job seekers from underrepresented groups to identify what "fair screening" means beyond demographic parity;  \textbf{Power-aware transparency}: Document and disclose whose definition of "qualified candidate" is encoded in the system—revealing whether criteria reflect genuine job requirements or historical biases of previous hiring managers, and how employer preferences versus candidate needs are balanced; \textbf{Context-sensitive trustworthiness}: Recognize that trustworthiness criteria differ by stakeholder—employers may prioritize efficiency and cultural fit, while candidates prioritize fairness and transparency in rejection reasons, and marginalized communities may prioritize systems that challenge rather than perpetuate historical exclusion patterns. \\
        \midrule
        Criminal risk assessment algorithms 
        & Evaluate prediction accuracy by measuring how well risk scores correlate with actual re-arrest or re-offense rates~\citep{dressel2018accuracy}; Conduct post-hoc bias and fairness audits when systems face legal challenges or public scrutiny~\citep{angwin2022machine}; 
        Measure system adoption rates and judicial satisfaction with decision support~\citep{stevenson2024algorithmic}. & \textbf{Differential impact evaluation}: Assess not just classification accuracy but downstream consequences across racial and socioeconomic groups—including cumulative disadvantage from pretrial detention, access to rehabilitation programs, employment prospects post-release, and intergenerational community impacts of mass incarceration; \textbf{Power-aware transparency}: Document and disclose whose conception of "risk" and "public safety" is embedded in the system—revealing whether models perpetuate historical policing patterns that over-surveilled marginalized communities, how feature selection reflects prosecutorial versus defense perspectives, and which stakeholder interests (law enforcement efficiency vs. defendant rights) dominate design decisions; \textbf{Redistributed evaluation authority}: Include criminal justice reform advocates, affected community members, defense attorneys, and social workers in defining evaluation success criteria alongside prosecutors. \\ 
        \midrule
        AI-powered adaptive learning platforms and automated assessment systems & 
        Assess prediction accuracy of student performance and dropout risk models; Measure engagement metrics (time-on-task, interaction frequency, content completion)~\citep{tan2025artificial}; 
        Conduct user satisfaction surveys with teachers and administrators~\citep{strielkowski2025ai}; Optimize for institutional metrics like pass rates and graduation rates~\citep{du2024personalized}. & \textbf{Longitudinal trust dynamics}: Track how student and teacher trust in the system evolves across academic failures or successes—designing trust repair mechanisms when the system provides ineffective learning paths or inaccurate assessments; \textbf{Differential impact evaluation}: Assess not just average learning gains but stratify outcomes across socioeconomic status, disability status, language background, and prior educational opportunity—revealing whether the system narrows or widens achievement gaps, and whether it channels certain populations toward vocational tracks while others receive enrichment; \textbf{Power-aware transparency}: Document and disclose whose pedagogical philosophy is embedded in the system—revealing whether learning objectives reflect standardized testing priorities versus holistic education; \textbf{Appropriateness of trust calibration}: Develop frameworks helping students and teachers understand when to trust AI recommendations (e.g., practice problem selection, pacing suggestions) versus when to override them (e.g., when algorithms misinterpret struggle as inability rather than productive challenge, or when standardized paths ignore individual passions and strengths).\\
        
\bottomrule
\end{tabularx}
\end{center}

Our proposed recommendations represent the first step rather than the end of achieving trustworthiness of AI systems in real-world applications. 
Realizing this vision requires sustained, collaborative efforts from researchers, practitioners, and stakeholders across multiple disciplines.

\subsection{Limitations}
First, our focus on AIES \& FAccT conferences reflects a particular scholarly perspective. To address this and build a more comprehensive understanding of how trustworthiness is conceptualized across AI research landscape, we propose extending our analysis to major technical AI conferences such as NeurIPS, ICML, IJCAI, and AAAI. This expanded corpus would enable comparison between how trustworthiness is framed in technically-focused venues versus interdisciplinary ones, potentially revealing important differences in priorities, metrics, evaluation methods, and theoretical frameworks. 

Second, our focus on academic discourse may not adequately represent how trustworthiness is understood by other key stakeholders, including industry practitioners, policymakers, and diverse user communities. These groups may prioritize different aspects of trustworthiness based on their  contexts and concerns.

Finally, cultural and geographical limitations exist in our analysis, as both AIES \& FAccT have historically featured stronger representation from Western institutions. This may result in overlooking important cultural variations in how trustworthiness is conceptualized across societies and knowledge traditions.

\section{Conclusion}
This study underscores the growing importance of trustworthiness in AI systems as a foundation for ethical and reliable technology. Through a review of AIES \& FAccT conference proceedings, we analyzed how trustworthiness is conceptualized, measured, and validated. While progress has been made in defining attributes such as transparency, accountability, and robustness, significant gaps remain, particularly in addressing the sociotechnical nature of AI and integrating safety into trustworthiness discussions.

Our findings highlight the need for an interdisciplinary approach that combines technical precision with social and ethical considerations. Current research often prioritizes technical attributes, overlooking the complex interplay between AI systems and the broader social, cultural, and institutional contexts in which they operate. Addressing these gaps is crucial for the AIES \& FAccT community to drive a paradigm shift, advancing holistic approaches to trustworthy AI that genuinely benefit society and promote responsible technological development.

\begin{acks}
    This research was (partially) supported by the Dutch Research Council (NWO), under project numbers 024.004.022, NWA.1389.20.\-183, and KICH3.LTP.20.006, and the European Union under grant agreements No. 101070212 (FINDHR) and No. 101201510 (UNITE).
    
    Views and opinions expressed are those of the author(s) only and do not necessarily reflect those of their respective employers, funders and/or granting authorities.
\end{acks}

\printbibliography

\clearpage
\appendix
\section*{Appendix}
\section{List of included articles in the final corpus for analysis}
\begin{enumerate}
\item \textbf{[FAccT]} Alpherts et al., Perceptive Visual Urban Analytics is Not (Yet) Suitable for Municipalities, \textit{Proceedings of the 2024 ACM Conference on Fairness, Accountability, and Transparency}.
\item \textbf{[FAccT]} Andrus et al., What We Can't Measure, We Can't Understand: Challenges to Demographic Data Procurement in the Pursuit of Fairness. \textit{Proceedings of the 2021 ACM Conference on Fairness, Accountability, and Transparency}.
\item \textbf{[FAccT]} Bhatt et al., Explainable machine learning in deployment. \textit{Proceedings of the 2020 Conference on Fairness, Accountability, and Transparency}.
\item \textbf{[FAccT]} Casper et al., Black-Box Access is Insufficient for Rigorous AI Audits. \textit{Proceedings of the 2024 ACM Conference on Fairness, Accountability, and Transparency}.
\item \textbf{[FAccT]} Engelmann et al., Clear Sanctions, Vague Rewards: How China's Social Credit System Currently Defines ``Good'' and ``Bad'' Behavior. \textit{Proceedings of the Conference on Fairness, Accountability, and Transparency, FAT* 2019}.
\item \textbf{[FAccT]} Engelmann et al., What People Think AI Should Infer From Faces. \textit{Proceedings of the 2022 ACM Conference on Fairness, Accountability, and Transparency}.
\item \textbf{[FAccT]} Ferrario et al., How Explainability Contributes to Trust in AI. \textit{Proceedings of the 2022 ACM Conference on Fairness, Accountability, and Transparency}.
\item \textbf{[FAccT]} Ferrario, A. A Trustworthiness-based Metaphysics of Artificial Intelligence Systems. \textit{Proceedings of the 2025 ACM Conference on Fairness, Accountability, and Transparency}.
\item \textbf{[FAccT]} Inie et al., From ``AI'' to Probabilistic Automation: How Does Anthropomorphization of Technical Systems Descriptions Influence Trust? \textit{Proceedings of the 2024 ACM Conference on Fairness, Accountability, and Transparency}.
\item \textbf{[FAccT]} Jacovi et al., Formalizing Trust in Artificial Intelligence: Prerequisites, Causes and Goals of Human Trust in AI. \textit{Proceedings of the 2021 ACM Conference on Fairness, Accountability, and Transparency}.
\item \textbf{[FAccT]} Kim et al., ``I'm Not Sure, But...'': Examining the Impact of Large Language Models' Uncertainty Expression on User Reliance and Trust. \textit{Proceedings of the 2024 ACM Conference on Fairness, Accountability, and Transparency}.
\item \textbf{[FAccT]} Kim et al., Humans, AI, and Context: Understanding End-Users Trust in a Real-World Computer Vision Application. \textit{Proceedings of the 2023 ACM Conference on Fairness, Accountability, and Transparency}.
\item \textbf{[FAccT]} Kinahan et al., Achieving Reproducibility in EEG-Based Machine Learning. \textit{Proceedings of the 2024 ACM Conference on Fairness, Accountability, and Transparency}.
\item \textbf{[FAccT]} Knowles et al., Trustworthy AI and the Logics of Intersectional Resistance. \textit{Proceedings of the 2023 ACM Conference on Fairness, Accountability, and Transparency}.
\item \textbf{[FAccT]} Knowles et al., The Sanction of Authority: Promoting Public Trust in AI. \textit{Proceedings of the 2021 ACM Conference on Fairness, Accountability, and Transparency}.
\item \textbf{[FAccT]} Liao et al., Designing for Responsible Trust in AI Systems: A Communication Perspective. \textit{Proceedings of the 2022 ACM Conference on Fairness, Accountability, and Transparency}.
\item \textbf{[FAccT]} Maeda et al., When Human-AI Interactions Become Parasocial: Agency and Anthropomorphism in Affective Design. \textit{Proceedings of the 2024 ACM Conference on Fairness, Accountability, and Transparency}.
\item \textbf{[FAccT]} Magaña et al., Frameworks, Methods and Shared Tasks: Connecting Participatory AI to Trustworthy AI Through a Systematic Review of Global Projects. \textit{Proceedings of the 2025 ACM Conference on Fairness, Accountability, and Transparency}.
\item \textbf{[FAccT]} Manzini et al., Should Users Trust Advanced AI Assistants? Justified Trust As a Function of Competence and Alignment. \textit{Proceedings of the 2024 ACM Conference on Fairness, Accountability, and Transparency}.
\item \textbf{[FAccT]} Nannini et al., Explainability in AI Policies: A Critical Review of Communications, Reports, Regulations, and Standards in the EU, US, and UK. \textit{Proceedings of the 2023 ACM Conference on Fairness, Accountability, and Transparency}.
\item \textbf{[FAccT]} Panigutti et al., The Role of Explainable AI in the Context of the AI Act. \textit{Proceedings of the 2023 ACM Conference on Fairness, Accountability, and Transparency}.
\item \textbf{[FAccT]} Paraschou et al., Ties of Trust: A Bowtie Model to Uncover Trustor-Trustee Relationships in LLMs. \textit{Proceedings of the 2025 ACM Conference on Fairness, Accountability, and Transparency}.
\item \textbf{[FAccT]} Pareek et al., Trust Development and Repair in AI-Assisted Decision-Making during Complementary Expertise. \textit{Proceedings of the 2024 ACM Conference on Fairness, Accountability, and Transparency}.
\item \textbf{[FAccT]} Scharowski et al., Certification Labels for Trustworthy AI: Insights From an Empirical Mixed-Method Study. \textit{Proceedings of the 2023 ACM Conference on Fairness, Accountability, and Transparency}.
\item \textbf{[FAccT]} Schoeffer et al., There Is Not Enough Information: On the Effects of Explanations on Perceptions of Informational Fairness and Trustworthiness in Automated Decision-Making. \textit{Proceedings of the 2022 ACM Conference on Fairness, Accountability, and Transparency}.
\item \textbf{[FAccT]} Shailya et al., LExT: Towards Evaluating Trustworthiness of Natural Language Explanations. \textit{Proceedings of the 2025 ACM Conference on Fairness, Accountability, and Transparency}.
\item \textbf{[FAccT]} Thornton et al., Fifty Shades of Grey: In Praise of a Nuanced Approach Towards Trustworthy Design. \textit{Proceedings of the 2021 ACM Conference on Fairness, Accountability, and Transparency}.
\item \textbf{[FAccT]} Thornton et al., The Alchemy of Trust: The Creative Act of Designing Trustworthy Socio-Technical Systems. \textit{Proceedings of the 2022 ACM Conference on Fairness, Accountability, and Transparency}.
\item \textbf{[FAccT]} Toney et al., Trust Issues: Discrepancies in Trustworthy AI Keywords Use in Policy and Research. \textit{Proceedings of the 2024 ACM Conference on Fairness, Accountability, and Transparency}.
\item \textbf{[FAccT]} Toreini et al., The Relationship between Trust in AI and Trustworthy Machine Learning Technologies. \textit{Proceedings of the 2020 Conference on Fairness, Accountability, and Transparency}.
\item \textbf{[FAccT]} Wang et al., Investigating and Designing for Trust in AI-powered Code Generation Tools. \textit{Proceedings of the 2024 ACM Conference on Fairness, Accountability, and Transparency}
\item \textbf{[AIES]} Brand et al., Exploring the Moral Value of Explainable Artificial Intelligence Through Public Service Postal Banks. \textit{Proceedings of the 2023 AAAI/ACM Conference on AI, Ethics, and Society}.
\item \textbf{[AIES]} Huang et al., Towards Just, Fair and Interpretable Methods for Judicial Subset Selection. \textit{Proceedings of the AAAI/ACM Conference on AI, Ethics, and Society}.
\item \textbf{[AIES]} Kim et al., Age Bias in Emotion Detection: An Analysis of Facial Emotion Recognition Performance on Young, Middle-Aged, and Older Adults. \textit{Proceedings of the 2021 AAAI/ACM Conference on AI, Ethics, and Society}.
\item \textbf{[AIES]} Lakkaraju et al., ``How do I fool you?'': Manipulating User Trust via Misleading Black Box Explanations. \textit{Proceedings of the AAAI/ACM Conference on AI, Ethics, and Society}.
\item \textbf{[AIES]} Lawrence et al., The Bureaucratic Challenge to AI Governance: An Empirical Assessment of Implementation at U.S. Federal Agencies. \textit{Proceedings of the 2023 AAAI/ACM Conference on AI, Ethics, and Society}.
\item \textbf{[AIES]} Loi et al., Towards Accountability in the Use of Artificial Intelligence for Public Administrations. \textit{Proceedings of the 2021 AAAI/ACM Conference on AI, Ethics, and Society}.
\item \textbf{[AIES]} Mohseni et al., Toward Design and Evaluation Framework for Interpretable Machine Learning Systems. \textit{Proceedings of the 2019 AAAI/ACM Conference on AI, Ethics, and Society}.
\item \textbf{[AIES]} Schmitz et al., Towards Formalizing and Assessing AI Fairness. \textit{Proceedings of the 2023 AAAI/ACM Conference on AI, Ethics, and Society}.
\item \textbf{[AIES]} Shelby et al., Sociotechnical Harms of Algorithmic Systems: Scoping a Taxonomy for Harm Reduction. \textit{Proceedings of the 2023 AAAI/ACM Conference on AI, Ethics, and Society}
\item \textbf{[AIES]} Williams et al., No Justice, No Robots: From the Dispositions of Policing to an Abolitionist Robotics. \textit{Proceedings of the 2023 AAAI/ACM Conference on AI, Ethics, and Society}.
\item \textbf{[AIES]} Zhang et al., U.S. Public Opinion on the Governance of Artificial Intelligence. \textit{Proceedings of the AAAI/ACM Conference on AI, Ethics, and Society}.
\item \textbf{[AIES]} Zilka et al., Transparency, Governance and Regulation of Algorithmic Tools Deployed in the Criminal Justice System: A UK Case Study. \textit{Proceedings of the 2022 AAAI/ACM Conference on AI, Ethics, and Society}.
\end{enumerate}

\section{Research Ethics and Social Impact}

\subsection{Ethical Considerations Statement}
This paper is primarily a literature review and theoretical contribution, thus we did not engage in any human subjects research, systems development or deployment. Our work has been guided by considerable efforts of the AIES \& FAccT community in the ethical ramifications of AI systems and their impact on human societies. This scoping review aims to contribute to a better understanding of AI trustworthiness, acknowledging the potential biases and limitations. We are committed to fostering responsible AI development and deployment, and this research is intended to support that goal.


\subsection{Adverse Impact Statement}
With this work, we seek to spark conversations across disciplines about the social and technical inequalities being seen in the literature while studying AI trustworthiness. Rather than ``calling out'' specific articles, we hope this critique serves as a ``call in'' other scholars to join together and work collectively towards critically understanding the trustworthiness of AI systems.

\subsection{Social Impact}
One of the primary impacts of AIES \& FAccT scholarship is the enhancement of public literacy regarding AI technologies. By critically examining the articles focusing on understanding trustworthiness of AI systems, we hope to have fostered a greater understanding of the complexities involved in trustworthy AI deployment, particularly concerning issues of comprehensive measurement frameworks, socio-technical integration and contribution to SDGs.


\section{Reproducibility Checklist for JAIR}

\subsection*{All articles:}

\begin{enumerate}
    \item All claims investigated in this work are clearly stated. 
    [yes]
    \item Clear explanations are given how the work reported substantiates the claims. 
    [yes]
    \item Limitations or technical assumptions are stated clearly and explicitly. 
    [yes]
    \item Conceptual outlines and/or pseudo-code descriptions of the AI methods introduced in this work are provided, and important implementation details are discussed. 
    [NA]
    \item 
    Motivation is provided for all design choices, including algorithms, implementation choices, parameters, data sets and experimental protocols beyond metrics.
    [yes]
\end{enumerate}

\end{document}